\documentclass{article}
\usepackage{amsmath,mlspconf}
\usepackage[utf8]{inputenc} 
\usepackage[T1]{fontenc}    
\usepackage{hyperref}       
\usepackage{url}            
\usepackage{booktabs}       
\usepackage{amsfonts}       
\usepackage{nicefrac}       
\usepackage{microtype}      
\usepackage{lipsum}
\usepackage{graphicx}
\usepackage{xcolor}
\usepackage[numbers]{natbib}
\usepackage{flushend}

\graphicspath{{figures/}}

\title{Hierarchical graph neural nets can capture long-range interactions}
\name{%
    Ladislav Rampášek\thanks{This work was partially funded by CIFAR AI Chair [\emph{G.W.}] and IVADO grant PRF-2019-3583139727. The content here is solely the responsibility of the authors and does not necessarily represent the official views of the funding agencies. Code: \url{https://github.com/rampasek/HGNet}. Correspondence to: \texttt{\{ladislav.rampasek,wolfguy\}@mila.quebec}}%
    \qquad Guy Wolf
}
\address{%
    Université de Montréal, Dept. of Math. \& Stat.; \ Mila - Quebec AI Institute, Montreal, QC, Canada
}

\begin{document}

\maketitle

\begin{abstract}
Graph neural networks (GNNs) based on message passing between neighboring nodes are known to be insufficient for capturing long-range interactions in graphs.
In this paper~we study hierarchical message passing models that leverage a multi-resolution representation of a given graph. This facilitates learning of features that span large receptive fields without loss of local information, an aspect not studied in preceding work on hierarchical GNNs. 
We introduce Hierarchical Graph Net (HGNet), which for any two connected nodes guarantees existence of message-passing paths of at most logarithmic length w.r.t.\ the input graph size. Yet, under mild assumptions, its internal hierarchy maintains asymptotic size equivalent to that of the input graph. We observe that our HGNet outperforms conventional stacking of GCN layers particularly in molecular property prediction benchmarks. Finally, we propose two benchmarking tasks designed to elucidate capability of GNNs to leverage long-range interactions in graphs.
\end{abstract}
%

\section{Introduction}
\label{sec:intro}
Graph neural networks (GNNs), and the field of geometric deep learning, have seen rapid development in recent years \cite{hamilton2020, bronstein2021G5} and have attained popularity in various fields involving graph and network structures. Prominent examples of GNN applications include molecular property prediction, physical systems simulation, combinatorial optimization, or interaction detection in images and text. Many of the current GNN designs are based on the principle of neural message-passing \cite{gilmer2017MP}, where information is iteratively passed between neighboring nodes along existing edges. However, this paradigm is known to suffer from several deficiencies, including theoretical limits of their representational capacity~\cite{xu2018gin} and observed limitations of their information propagation over graphs~\cite{alon2020bottleneck,li2018insights,min2020scattering}. 

Two of the most prominent deficiencies of GNNs are known as \emph{oversquashing} and \emph{oversmoothing}. Information \emph{oversquashing} refers to the exponential growth in the amount of information that has to be encoded by the network with each message-passing iteration, which rapidly grows beyond the capacity of a fixed hidden-layer representation~\cite{alon2020bottleneck}. Signal \emph{oversmoothing} refers to the tendency of node representations to converge to local averages~\cite{li2018insights}, which can also be observed in graph convolutional networks implementing low pass filtering over the graph~\cite{min2020scattering}. A significant repercussion of these phenomena is that they limit the ability of most GNN architectures to represent long-range interactions (LRIs) in graphs. Namely, they struggle in capturing dependencies between distant nodes, even when these have potentially significant impact on output prediction or appropriate internal feature extraction towards it. Capturing LRIs typically requires the number of GNN layers (i.e., implementing individual message passing steps) to be proportional to the diameter of the graph, which in turn exacerbates the oversquashing of massive amount of information and the oversmoothing that tends towards averaging over wide regions of the graph, if not the entire graph.

In this paper, we study the utilization of multiscale hierarchical meta-structures to enhance message passing in GNNs and facilitate capturing of LRIs. By leveraging hierarchical message passing between nodes, our Hierarchical Graph Net (HGNet) architecture can propagate information within $O(\log |V(G)|)$ steps instead of $O(\mathrm{diam}(G))$, leading to particular improvements for sparse graphs with large diameters. 

We note that a few works have recently proposed related approaches using hierarchical constructions, namely g-U-Net \cite{gao2019gUnet} and GXN \cite{li2020GXN}. g-U-Net employs a similarity-based top-k pooling called gPool for hierarchical construction over which it implements bottom-up and simple top-down message passing. GXN introduced mutual information based pooling (VIPool) together with a more complex cross-level message passing.
Next, MGKN \cite{li2020multipole} introduced  multi-resolution GNN with V-cycle algorithm specifically for learning solutions operators to PDEs.
Broadly related are also differentiable pooling methods such as DiffPool~\cite{ying2018diffpool}, EdgePool~\cite{diehl2019edgepool}, or GraphZoom~\cite{deng2020graphzoom}. However, these do not employ two-directional hierarchical message passing.


While LRIs are widely accepted as being important for both theoretical studies and in practice, most benchmarks used to empirically validate GNN models do not clearly exhibit this property. Out of these, the importance of LRIs is perhaps best justified in biochemistry datasets, where the 2D structure of proteins and molecules is  used as their graph representation. However, edges of such graphs do not encode 3D forces and global properties, leaving it up to the model to learn to recognize such LRIs. Several highly specialized models have been proposed for molecular data, but these are typically not applicable to other domains, which also hinders analysis of their modeling improvements towards particularly capturing LRIs. Therefore, in our experiments we primarily focus on quantifying the benefit of using a hierarchical structure compared to the standard practice of GNN layer stacking. We also introduce two benchmarking tasks designed to elucidate capability of general-purpose GNNs to leverage LRIs. Here, we show hierarchical models outperform their standard GNN counterparts when their hierarchical graph construction matches well with the original graph structure and the prediction task, while uncovering related limitations of gPool in g-U-Net.

\section{Hierarchical graph net}
\label{sec:hgnet}
To build a hierarchical message passing model, we need to construct a hierarchical graph representation and define an inter- and intra-level message passing mechanism.

\subsection{Graph coarsening for hierarchical representation}
Building a hierarchical representation principally involves iterative application of graph coarsening and pooling operations. Graph coarsening computes a mapping from nodes of a starting graph $G_\ell$ onto nodes of a new smaller graph $G_{\ell+1}$, while the pooling step computes node and edge features of $G_{\ell+1}$ from $G_\ell$. Here we explore two different approaches: EdgePool~\cite{diehl2019edgepool} and the Louvain method for community detection~\cite{blondel2008Louvain}.

\textbf{EdgePool} \cite{diehl2019edgepool} is a method based on the principle of edge contractions. First, the raw score of an edge $e_{u,v}=(u, v)$ is obtained by a linear combination of respective node features $x_u$ and $x_v$: $r_{u,v} = W (x_u || x_v) + b$. Raw scores of edges incident to a node $u$ are then normalized as $0.5 + \mathrm{softmax}_{v\in \mathcal{N}(u)} r_{u,v}$ to obtain the final edge scores $s_{u,v}$. Finally, a maximal set of edges is greedily selected according to their scores $s_{a,b}$ and then contracted to create a new graph $G_{\ell+1}$ from $G_\ell$, while nodes in $G_\ell$ that were not merged are carried forward to $G_{\ell+1}$. Two nodes $a, b$ in $G_{\ell+1}$ are then connected by an edge iff there exist two nodes in $G_\ell$ the $a, b$ were constructed from that had been adjacent in $G_\ell$.

Contraction of an edge $(u^{(\ell)},v^{(\ell)}) \in G_\ell$ results in a new node $w^{(\ell+1)}$ with features $x^{(\ell+1)}_w := s^{(\ell)}_{u,v} (x^{(\ell)}_u + x^{(\ell)}_v)$. Multiplying the new node features by the edge score facilitates gradient-based learning of the scoring function, which would otherwise be independent of the final objective function.





\textbf{Louvain method} for community detection \cite{blondel2008Louvain} is a heuristic method based on greedy maximization of modularity score of each community. It is an $O(N \log N)$ algorithm without learnable parameters that is deterministic for a fixed random seed. The Louvain algorithm merges clusters (communities) into a single node and iteratively performs modularity clustering on the condensed graph until the score cannot be improved. The size of the condensed graph cannot be directly controlled, but seems to yield satisfying contraction ratios in practice.

To build a hierarchical meta-graph over a starting graph $G_0$, we use average node and edge feature pooling according to the modular communities identified in $G_\ell$ by the Louvain method to construct the following level $G_{\ell+1}$.


\begin{figure}[tb]
\begin{minipage}[]{1.0\linewidth}
  \centering
  \centerline{\includegraphics[width=8.7cm]{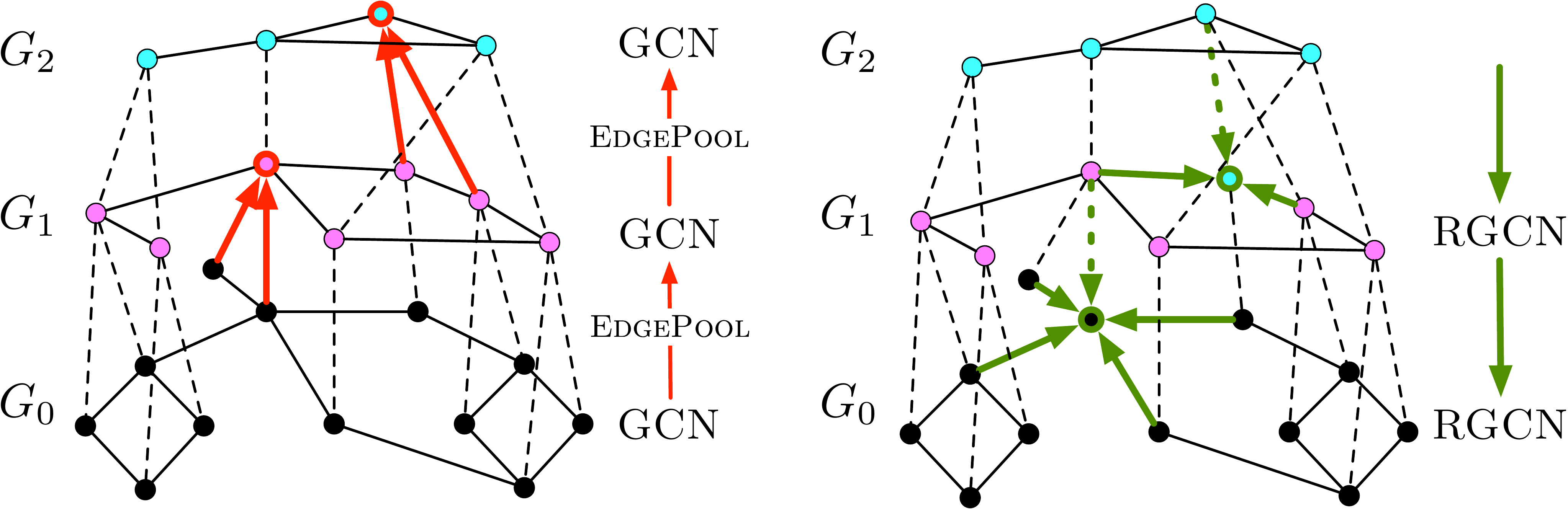}}
\end{minipage}
\caption{\textbf{HGNet} with two hierarchical levels over an original graph $G_0$ of 12 vertices (in black) and 14 edges. The dashed lines represent inter-level edges. (left) Two levels of EdgePool coarsening, highlighted by red arrows, create the hierarchical structure. A GCN layer is applied before each EdgePool coarsening and at the final coarsest level $G_2$. (right) Message passing down the hierarchy is implemented by an RGCN layer at $G_1$ and then $G_0$ levels, highlighted by green arrows, where inter-level edges are treated as a distinct edge type.}
\label{fig:hgnet}
\end{figure}

\subsection{Hierarchical message passing in HGNet}
Both EdgePool and the Louvain method provide a recipe for construction of a hierarchical graph representation. We propose Hierarchical Graph Network (HGNet) based on either one of these approaches (see Figure \ref{fig:hgnet}), sharing the same hierarchical message passing approach that we describe next. Our message passing both within and between levels is principally similar to that of g-U-Net. Consider a hierarchical meta-graph with $L$ levels over some $G_0$. The forward propagation in HGNet consists of a computational pass going up the hierarchy and of a pass going down the hierarchy, resulting in the final embedding of each node in $G_0$. In the upwards pass we first apply a GCN layer \cite{kipf2016GCN} to $G_\ell$, starting with $\ell=0$, followed by node and edge pooling according to either EdgePool or the Louvain method to instantiate the next hierarchical level $G_{\ell+1}$. This process iterates until the final level $L$, at which point no more pooling is done and the downwards pass starts. In this downwards pass we utilize RGCN \cite{schlichtkrull2018RGCN} layers at each $G_\ell$ level $\ell \in \{L-1, \dots, 0\}$, where we add special edges that connect merged nodes in $G_\ell$ with their respective representatives in $G_{\ell+1}$ by an edge of unique type.

\textbf{Complexity.} We now analyze the asymptotic complexity of our hierarchical meta-graph based on the EdgePool variant. 
Let us assume that in each round of edge contractions the size of the greedy maximum matching is at least a constant fraction $m \geq 2$ of the number of remaining nodes, i.e., $\frac{N}{m}$. Note that $m=2$ when the selected set of edges is a perfect matching. That means after the first round there will be $N\frac{m-1}{m}$ nodes in the next $G_1$ level. Thus, the total number of nodes in the entire hierarchical structure over a $G_0$ with $N$ nodes is $\sum_{\ell=0}^\infty N \left(\frac{m-1}{m}\right)^\ell = mN = O(N)$, while the number of possible levels is $\log_{\frac{m}{m-1}} N = O(\log N)$. This construction therefore guarantees that, if $G_0$ is connected, the shortest path length between any two nodes is upper-bounded by $O(\log N)$. 

We can also expect the number of edges in our hierarchical graph to remain asymptotically equal to the number of edges in the input graph $G_0$.
Assume there are $E = \Omega(N)$ edges in $G_0$ out of $O(N^2)$ possible and that they are uniformly distributed. Then after one round of EdgePool, the number of edges in $G_1$ is expected to be $O(E\left(\frac{m-1}{m}\right)^2)$, because the number of possible edges in $G_1$ compared to $G_0$ has decreased from $O(N^2)$ to $O(\left[N\frac{m-1}{m}\right]^2)$, i.e., we can expect $\left(\frac{m-1}{m}\right)^2$ contraction factor for the number of edges. Therefore, we can expect
$\sum_{\ell=0}^\infty E \left(\frac{m-1}{m}\right)^{2\ell} = \frac{m^2}{2m-1}E = O(E)$
intra-level edges in total. From the construction of the hierarchy it is also clear that the number of inter-level edges (connecting nodes between adjacent hierarchical levels) is $O(N)$ as the total number of nodes is $O(N)$. Therefore, the total number of edges is expected to remain $O(E)$. 

Given a deep enough hierarchy and large enough node representation capacity, the final node embeddings can incorporate LRIs from the entire graph $G_0$, as well as local information. In the case of EdgePool, the asymptotic complexity of our HGNet remains that of GCN, as even despite our hierarchical graph having up to $O(\log N)$ hierarchical levels, its size remains asymptotically unchanged under reasonable assumptions. For a standard message passing GNN to theoretically achieve this capability, it is necessary to stack $O(\mathrm{diam}(G))$ layers, which may be prohibitively expensive.





\section{Results}
\label{sec:results}

\begin{table*}[htb]
\caption{\textbf{Legacy graph benchmarks.} CiteSeer, Cora and PubMed provide only one standard data split, and therefore we show test accuracy averaged over three runs with different random seeds for these datasets. For graph classification tasks (right side of the table) we used 10-fold stratified cross-validation. Shown heatmaps are normalized per dataset (column). 
}
\vspace{2pt}
\centering
\includegraphics[width=0.95\linewidth]{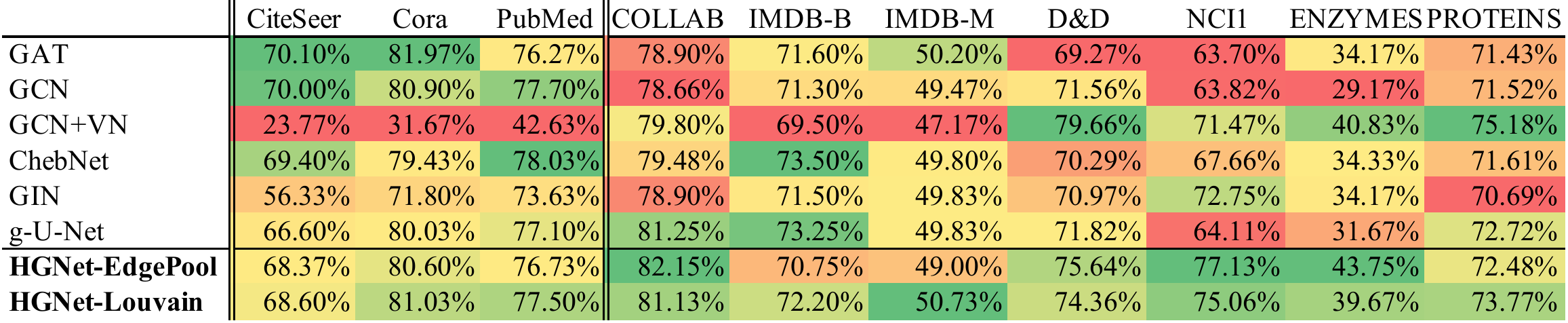}
\label{tab:res-std}
\end{table*}

In order to evaluate the performance of HGNet, we consider a wide variety of graph data, including transductive node classification and inductive graph-level classification. Our benchmarks include two settings of HGNet (namely, with EdgePool and Louvain hierarchical structures) and six competitive baseline models: GCN~\cite{kipf2016GCN}, GCN+VN (GCN extended with a Virtual Node connected to all other nodes), GAT~\cite{velickovic2017GAT}, ChebNet~\cite{tang2019chebnet}, GIN~\cite{xu2018gin}, and g-U-Net~\cite{gao2019gUnet}. The experimental setup is identical for all tested methods. Each method is trained for 200 epochs, followed by a selection of the best model based on the validation performance, and finally performance on the test split is reported. In case of GCN, GCN+VN, GAT, ChebNet and GIN, we always used a stack of 2 layers unless explicitly stated otherwise. In the case of g-U-Net, we reproduced published hyperparameters~\cite{gao2019gUnet} as closely as possible. For each method we default to 32-dimensional hidden node representation; other hyperparameters specific to certain tasks or datasets are described in the respective sections. We note that our reproduced g-U-Net results differ from the original publication~\cite{gao2019gUnet}, as there only the best validation set results were reported rather than performance on independent test sets. This erroneous practice had occurred on several occasions in the relatively nascent field of graph deep learning~\cite{errica2019fair}.

\subsection{Node classification in citation networks}

For our first benchmark, we consider semi-supervised node classification on the CiteSeer, Cora and PubMed citation networks~\cite{yang2016planetoid}. Our HGNet variants are configured with one hierarchical level and g-U-Net with four levels as per published hyperparameters. Citation networks are known to exhibit high homophily~\cite{zhu2020BeyondHomophily}, i.e., nodes tend to have the same class label as most of their first degree neighbors. First-order message passing GNNs are known to perform well in high-homophily settings \cite{zhu2020BeyondHomophily}, which is validated by our experiments presented in Table \ref{tab:res-std}, with the exception of GCN+VN and GIN. All three hierarchical methods (i.e., g-U-Net, HGNet-EdgePool, and HGNet-Louvain) attain very similar results, slightly behind the best performing GAT, GCN, and ChebNet. 

The low performance of GCN+VN, a model geared towards capturing global information, and middle-of-the-pack performances of the hierarchical methods can be explained by the high homophily present in the data, and support prior findings \cite{huang2020lp} showcasing that global graph information is not vital in these datasets. Hence, given similar model capacity and experimental settings, methods favoring local information, such as GAT and GCN, outperform the more sophisticated ones. We conclude that CiteSeer, Cora and PubMed are not directly suitable to test the ability of GNN models to capture global information or LRIs, despite their extensive use and popularity in such benchmarks \cite{gao2019gUnet, li2020GXN}.


\begin{table}[htb]
\caption{\textbf{Citation networks with $k$-hop sanitized dataset splits.} The reported metric is the average test accuracy over three training runs with different random seeds, while keeping the same resampled splits. Heatmaps are normalized per block given by a dataset and neighborhood size $k$ combination.
}
\vspace{2pt}
\begin{minipage}[]{1.0\linewidth}
  \centering
  \centerline{\includegraphics[width=8.5cm]{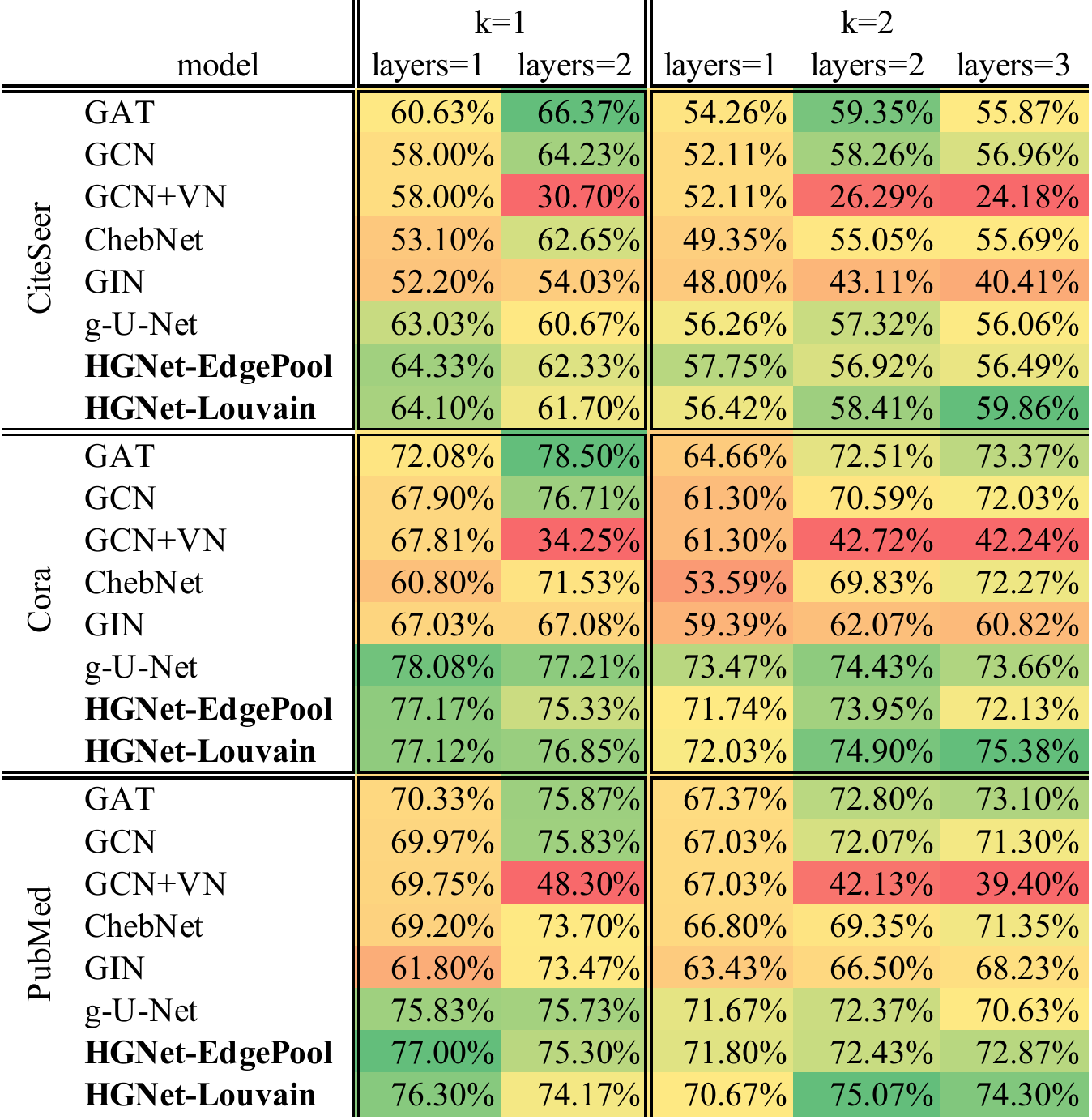}}
\end{minipage}
\label{tab:res-sanitmod}
\end{table}

\subsubsection*{Resampled citation networks} In an effort to make the prediction tasks of CiteSeer, Cora and PubMed citation networks more suitable for testing the models' ability to utilize information from farther nodes, we experimented with a specific resampling of their training, validation and test splits. The standard semi-supervised splits \cite{yang2016planetoid} follow the same key for each dataset: 20 examples from each class are randomly selected for training, while 500 and 1000 examples are drawn uniformly randomly for the validation and test splits. We used principally the same key, but a different random sampling strategy. Once a node is drawn, we enforced that none of its $k$-th degree neighbors is selected for any split. This approach guarantees that a $k$-hop neighborhood of each labeled node is ``sanitized'' of labels. As such, we prevent potential correct-class label imprinting in the representation of these $k$-th degree neighbors during the semi-supervised transductive training. For a model to leverage such imprinting benefit of homophily, it has to be able to reach beyond this $k$-hop neighborhood, assuming that the class homophily spans that far in the underlying data.

We experimented with $k\in\{1,2\}$ for all 3 citation networks and kept the same hyperparameters from the prior experiments, but varied the number of stacked layers or hierarchy levels, as applicable, for each GNN method. Results averaged over runs with 3 random seeds are shown in Table \ref{tab:res-sanitmod}. For $k=1$ we see consistent degradation of performance for single-layer GNNs, while even one level of hierarchy provides significant advantage for the hierarchical models. GAT and GCN recover competitive performance given two layers, which allows the models to reach second-order neighborhood with some nodes that are labeled during training. Hierarchical models however do not benefit from using two levels, as with even just one level their receptive field is already large enough to reach beyond first-order neighborhood of a node. In case of $k=2$ we observe similar behavior, but now hierarchical models typically benefit from employing two or three levels. This is particularly true for PubMed, the largest tested dataset. In this scenario we believe we have reached the limit of these datasets in the sense that we do not expect third-degree or further nodes to be consistently of significant relevance. We can see that for most methods the performance is relatively similar between two or three layers. Our resampling approach is fundamentally limited by the strong local homophily present in these citation networks and beyond $k=2$ cannot be used to test capability of the models to leverage LRIs.

\subsection{Graph-level prediction}

\begin{table*}[htb]
\caption{\textbf{OGB molecular benchmarks.} HGNet results are obtained and presented as per OGB standards, shown is the mean and standard deviation from 10 runs with different random seeds. HGNet models have 1, 2, or 3 levels and otherwise mirror hyperparameters of the OGB baselines that each have 5 layers. The metrics for baselines are from the OGB online leaderboard.}
\vspace{5pt}
\centering
\includegraphics[width=0.92\linewidth]{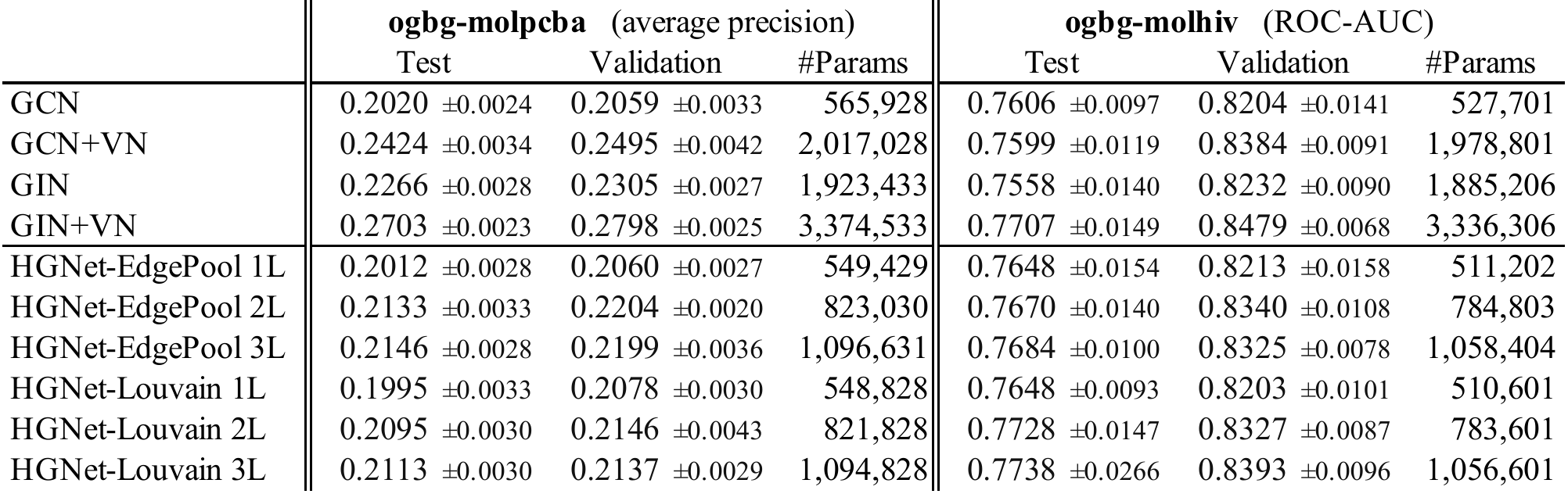}
\label{tab:res-ogbgmol}
\end{table*}

\begin{table*}[htb]
\caption{\textbf{Color-connectivity datasets.} The average test accuracy in 10-fold stratified CV for various depths of the models.}
\vspace{5pt}
\begin{minipage}[]{1.0\linewidth}
  \centering
  \centerline{\includegraphics[width=18cm]{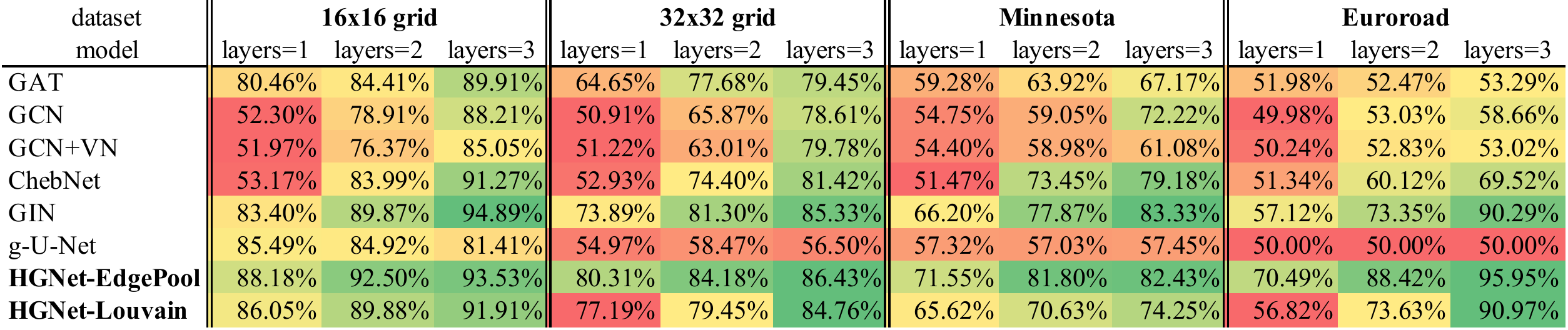}}
\end{minipage}
\label{tab:res-islands}
\end{table*}


We now turn our focus to graph-level classification. We start by benchmarking all methods using a set of commonly used datasets: COLLAB, IMDB-BINARY, IMDB-MULTI, D\&D, NCI1, ENZYMES, and PROTEINS \cite{morris2020tudataset}. In the second part we present a new set of datasets we designed to challenge the GNN methods in learning to recognize a complex set of features. In this section, we use global mean pooling for each method to obtain the graph-level representation from individual nodes of a graph. Using this representation, a graph is finally classified by a 2-layer MLP classifier with 128-dimensional hidden layer.

Our experimental results in common graph-classification datasets are presented in Table \ref{tab:res-std} (right side). One of our HGNet variants is the best performing method in 4 out of the 7 datasets. GCN+VN performs well on molecular datasets where global information is important, as does HGNet. However, g-U-Net falls behind in this setting, likely due to the nature of top-k pooling in its gPool, which destroys local information and appears to have difficulty extracting complex global features.

\subsubsection*{OGB molecular benchmarks} 
We tested HGNet on two Open Graph Benchmark (OGB) \cite{hu2020OGB} molecular property prediction datasets: \textit{ogbg-molpcba} and \textit{ogbg-molhiv}. For our HGNet we used the same experimental setup and GCN layer implementation as provided by OGB. Both EdgePool and Louvain versions of HGNet with 2 hierarchical levels (2L), composed of 3 GCN and 2 RGCN-like layers, outperform GCN with 5 layers (see Table~\ref{tab:res-ogbgmol}). Employing a hierarchical meta-graph is more powerful than stacking the same number of layers. We note that adding global readouts via Virtual Node is remarkably beneficial in \textit{ogbg-molpcba}, albeit at the cost of many additional parameters.

\subsubsection*{Color-connectivity task}
Open Graph Benchmark and other recent initiatives are increasing the bar for GNN benchmarking, as many established benchmarking datasets are too small or too simple to adequately test the expressive power of new GNN methods. However, the motivation to include a new dataset in a suite is typically based on the interest in a particular application domain and the scale of the dataset. Unfortunately, none of the existing benchmarks provably require the capture of LRIs for significant performance gain. This issue was not realized in the benchmarking of prior hierarchical methods \cite{gao2019gUnet, li2020GXN}, except \cite{stachenfeld2020SMP} that proposed shortest path prediction task in random graphs. Here we propose to employ a task not used for GNN benchmarking before -- classifying the connectivity of same colored nodes in graphs of varying topology. Our color-connectivity datasets are created by taking a graph and randomly coloring half of its nodes one color, e.g., red, and the other nodes blue, such that the red nodes either create a single connected island or two disjoint islands. The binary classification task is then distinguishing between these two cases. The node colorings were sampled by running two red-coloring random walks starting from two random nodes. We used 16x16 and 32x32 2D grids, as well as the Euroroad and Minnesota road networks \cite{rossi2015NR} for the underlying graph topology. For each, we sampled a balanced set of 15,000 examples, except for Minnesota network for which we generated 6,000 examples due to memory constraints. Solving this task requires combination of local and long-range information, while a global readout, e.g., via Virtual Node, is expected to be unsatisfactory.

HGNet-EdgePool is the single best method in this suite of benchmarks (Table \ref{tab:res-islands}). Given the nature of the data, we observe a large difference in how suitable are the hierarchical graphs created by different approaches. In particular, gPool of g-U-Net fails to facilitate the learning process on large graphs. Next, global readout via Virtual Nodes in the GCN+VN model does not provide any improvement over the standard GCN, as evidently it is not able to capture complex features. On the other hand, we see that the ChebNet and GIN models perform well. ChebNet can learn filters that have large receptive field in graph space, which is important in this case. We suspect that GIN is powerful enough to learn local heuristics GCN and GAT fail to, which warrants further investigation. 
\section{Conclusion}
Across many datasets, we saw hierarchical models outperform their standard GNN counterparts when construction of the hierarchical graph (its inductive bias) matches well with the graph structure and prediction task. 
We have not compared to methods highly specialized for a particular tasks, e.g., molecular property prediction, but rather focused on elucidating the effect of using a hierarchical structure compared to the standard approach of stacking GNN layers. Further research remains to be done in terms of exploring combinations of various pooling approaches, hierarchical message passing algorithms and utilization of, e.g., GIN layers instead of GCN.
Our proposed color-connectivity task requires complex graph processing to which most existing message-passing GNNs do not scale. These datasets can serve as a common-sense validation for new and more powerful methods. Our testbed datasets can still be improved, as the node features are minimal and recognition of particular topological patterns (e.g., rings or other subgraphs) is not needed to solve the current task.
Nevertheless, it represents a significant step forward in terms of understanding and benchmarking more complex graph neural networks.

\vspace{-5pt}
\paragraph*{Acknowledgments:} The authors would like to thank William L. Hamilton for insightful discussions and Semih Cantürk for help with proofreading of the manuscript.

\bibliographystyle{IEEEbib}
\vspace{-5pt}
\bibliography{references}

\appendix
\newcommand\Tstrut{\rule{0pt}{4ex}}         
\newcommand\Bstrut{\rule[-2ex]{0pt}{0pt}}   

\section*{Appendix}
\renewcommand{\thesubsection}{A.\arabic{subsection}}
\setcounter{figure}{0}
\renewcommand\thefigure{A.\arabic{figure}}  
\setcounter{table}{0}
\renewcommand\thetable{A.\arabic{table}}  




\begin{table*}[htb]
\caption{\textbf{Datasets summary.} The graph statistics are computed over all graphs in respective datasets. For model evaluation we used either the standard train/validation/test split as provided with the respective benchmark dataset and repeated the experiment 10 times with different random seeds (10x RS standard split); or we used 10-fold stratified cross-validation protocol (10-fold stratified CV). In transductive semi-supervised node classification with $k$-hop sanitized node pre-filtering we followed the same train/validation/test splitting procedure of \cite{yang2016planetoid}; train: 20 random pre-filtered nodes per each class, validation: 500 random nodes from remaining pre-filtered nodes, and test: 1000 random nodes from remaining pre-filtered nodes. }
\vspace{8pt}
\centering
\begin{tabular}{llllllrr}
\hline
\multicolumn{1}{c}{\textbf{Dataset}} & \multicolumn{1}{c}{\textbf{\#   Graphs}} & \multicolumn{1}{c}{\textbf{\begin{tabular}[c]{@{}c@{}}(avg.)\\      \# Nodes\end{tabular}}} & \multicolumn{1}{c}{\textbf{\begin{tabular}[c]{@{}c@{}}(avg.)\\      \# Edges\end{tabular}}} & \multicolumn{1}{c}{\textbf{\begin{tabular}[c]{@{}c@{}}(Node)\\      Features\end{tabular}}} & \multicolumn{1}{c}{\textbf{\#   Classes}}                                                  & \multicolumn{1}{c}{\textbf{Evaluation}} & \multicolumn{1}{c}{\textbf{Metric}} \Tstrut\Bstrut\\
\hline 
Cora                                 & 1                                        & 2,708                                                                                       & 5,429                                                                                       & 1,433                                                                                       & 7                                                                                          & 10x   RS standard split                 & accuracy                            \Tstrut\\
CiteSeer                             & 1                                        & 3,327                                                                                       & 4,552                                                                                       & 3,703                                                                                       & 6                                                                                          & 10x   RS standard split                 & accuracy                            \ \\
PubMed                               & 1                                        & 19,717                                                                                      & 44,338                                                                                      & 500                                                                                         & 3                                                                                          & 10x   RS standard split                 & accuracy                            \Bstrut\\
\hline
COLLAB                               & 5,000                                    & 74.49                                                                                       & 2457.78                                                                                     & \multicolumn{1}{c}{\begin{tabular}[c]{@{}c@{}}node\\      degree\end{tabular}}              & 3                                                                                          & 10-fold   stratified CV                 & accuracy                            \Tstrut\Bstrut\\
IMDB-BINARY                          & 1,000                                    & 19.77                                                                                       & 96.53                                                                                       & \multicolumn{1}{c}{\begin{tabular}[c]{@{}c@{}}node\\      degree\end{tabular}}              & 2                                                                                          & 10-fold   stratified CV                 & accuracy                            \Tstrut\Bstrut\\
IMDB-MULTI                           & 1,500                                    & 13                                                                                          & 65.94                                                                                       & \multicolumn{1}{c}{\begin{tabular}[c]{@{}c@{}}node\\      degree\end{tabular}}              & 3                                                                                          & 10-fold   stratified CV                 & accuracy                            \Tstrut\Bstrut\\
D\&D                                 & 1,178                                    & 284.32                                                                                      & 715.66                                                                                      & 89                                                                                          & 2                                                                                          & 10-fold   stratified CV                 & accuracy                            \Tstrut\\
NCI1                                 & 4,110                                    & 29.87                                                                                       & 32.3                                                                                        & 37                                                                                          & 2                                                                                          & 10-fold   stratified CV                 & accuracy                           \ \\
ENZYMES                              & 600                                      & 32.63                                                                                       & 62.14                                                                                       & 3                                                                                           & 6                                                                                          & 10-fold   stratified CV                 & accuracy                            \ \\
PROTEINS                             & 1,113                                    & 39.06                                                                                       & 72.82                                                                                       & 3                                                                                           & 2                                                                                          & 10-fold   stratified CV                 & accuracy                            \Bstrut\\
\hline
ogbg-molpcba                         & 437,929                                  & 26                                                                                          & 28.1                                                                                        & \multicolumn{1}{r}{\begin{tabular}[c]{@{}r@{}}9   node f.\\      3 edge f.\end{tabular}}    & \multicolumn{1}{r}{\begin{tabular}[c]{@{}r@{}}128   binary\\      multilabel\end{tabular}} & 10x   RS standard split                 & avg.   precision                    \Tstrut\Bstrut\\
ogbg-molhiv                          & 41,127                                   & 25.5                                                                                        & 27.5                                                                                        & \multicolumn{1}{r}{\begin{tabular}[c]{@{}r@{}}9   node f.\\      3 edge f.\end{tabular}}    & 2                                                                                          & 10x   RS standard split                 & ROC-AUC                             \Tstrut\Bstrut\\
\hline
C-C 16x16 grid                         & 15,000                                   & 256                                                                                         & 480                                                                                         & 1                                                                                           & 2                                                                                          & 10-fold   stratified CV                 & accuracy                            \Tstrut\\
C-C 32x32 grid                         & 15,000                                   & 1,024                                                                                       & 1,984                                                                                       & 1                                                                                           & 2                                                                                          & 10-fold   stratified CV                 & accuracy                            \ \\
C-C Euroroad                             & 15,000                                   & 1,174                                                                                       & 1,417                                                                                       & 1                                                                                           & 2                                                                                          & 10-fold   stratified CV                 & accuracy                           \ \\
C-C Minnesota                            & 6,000                                    & 2,642                                                                                       & 3,304                                                                                       & 1                                                                                           & 2                                                                                          & 10-fold   stratified CV                 & accuracy          \Bstrut\\
\hline
\end{tabular}
\label{tab:ds-stats}
\end{table*}

\begin{figure}[htb]
\begin{minipage}[b]{0.48\linewidth}
  \centering
  \centerline{\includegraphics[width=4.0cm]{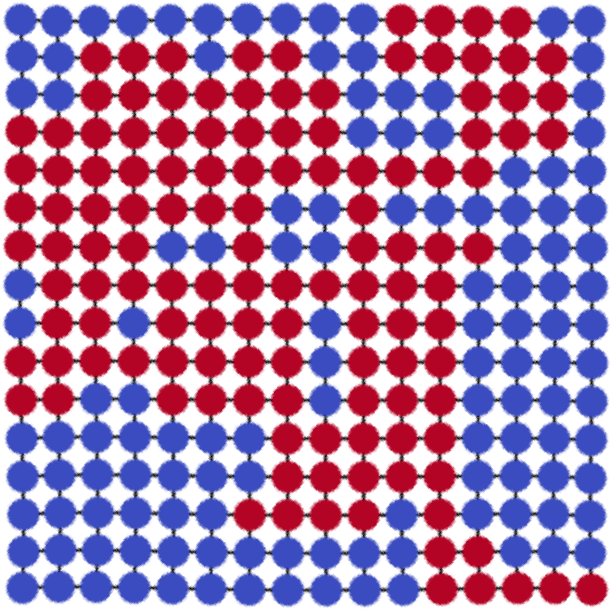}}
  \centerline{(a) label = 0}\medskip
\end{minipage}
\hfill
\begin{minipage}[b]{.48\linewidth}
  \centering
  \centerline{\includegraphics[width=4.0cm]{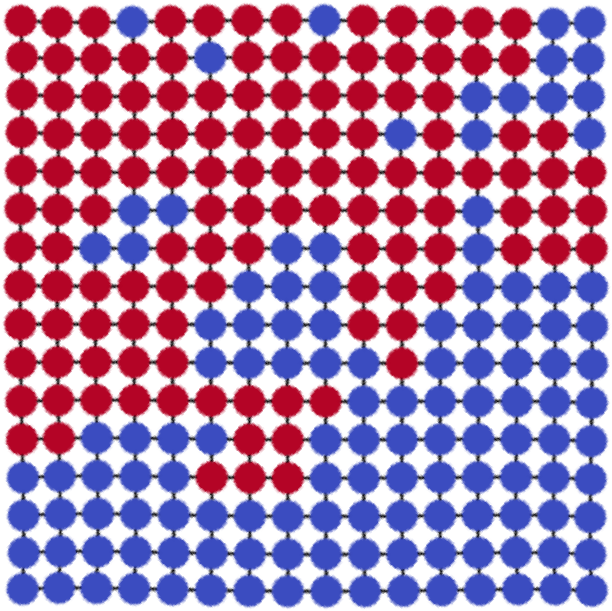}}
  \centerline{(b) label = 0}\medskip
\end{minipage}
\begin{minipage}[b]{0.48\linewidth}
  \centering
  \centerline{\includegraphics[width=4.0cm]{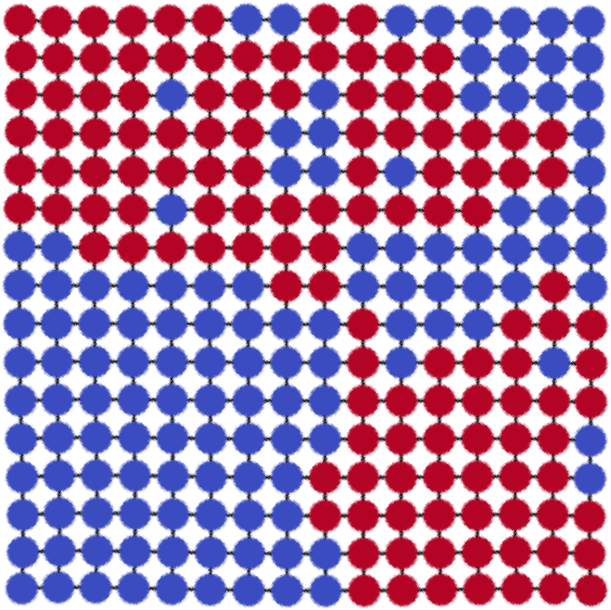}}
  \centerline{(c) label = 1}\medskip
\end{minipage}
\hfill
\begin{minipage}[b]{0.48\linewidth}
  \centering
  \centerline{\includegraphics[width=4.0cm]{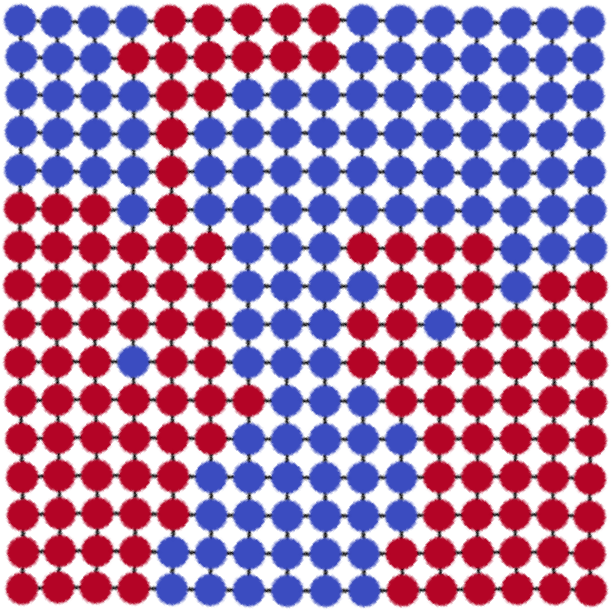}}
  \centerline{(d) label = 1}\medskip
\end{minipage}
\caption{Two negative and two positive examples from 16x16 grid Color-connectivity dataset.}
\label{fig:islands16x16viz}
\end{figure}

\begin{figure}[htb]
\begin{minipage}[b]{0.48\linewidth}
  \centering
  \centerline{\includegraphics[width=4.0cm]{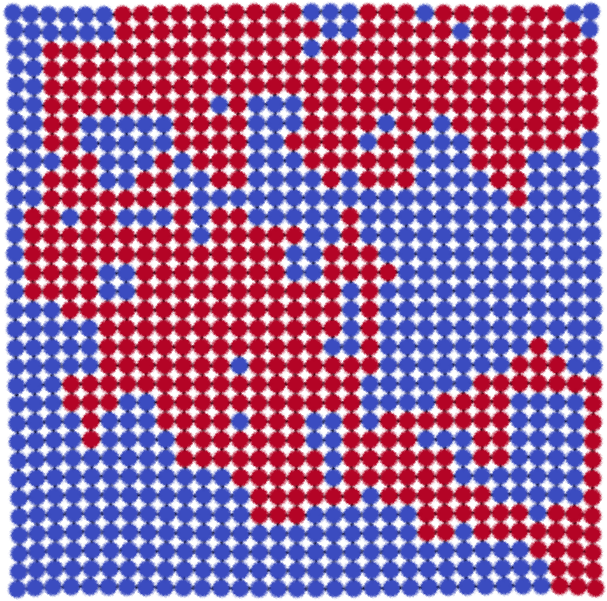}}
  \centerline{(a) label = 0}\medskip
\end{minipage}
\hfill
\begin{minipage}[b]{.48\linewidth}
  \centering
  \centerline{\includegraphics[width=4.0cm]{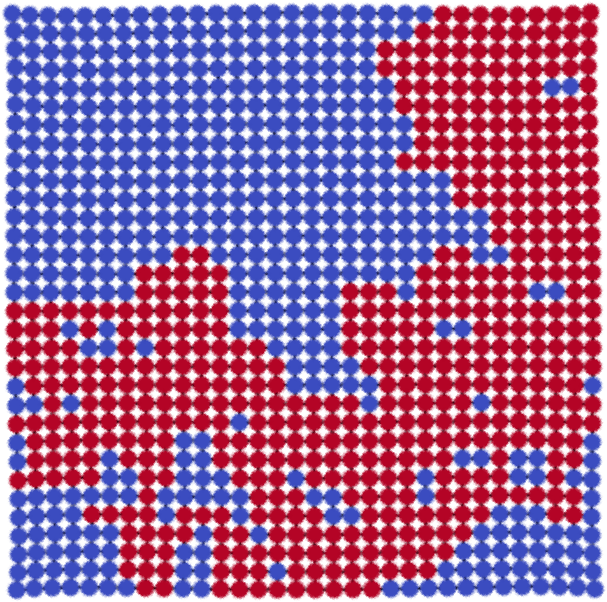}}
  \centerline{(b) label = 0}\medskip
\end{minipage}
\begin{minipage}[b]{0.48\linewidth}
  \centering
  \centerline{\includegraphics[width=4.0cm]{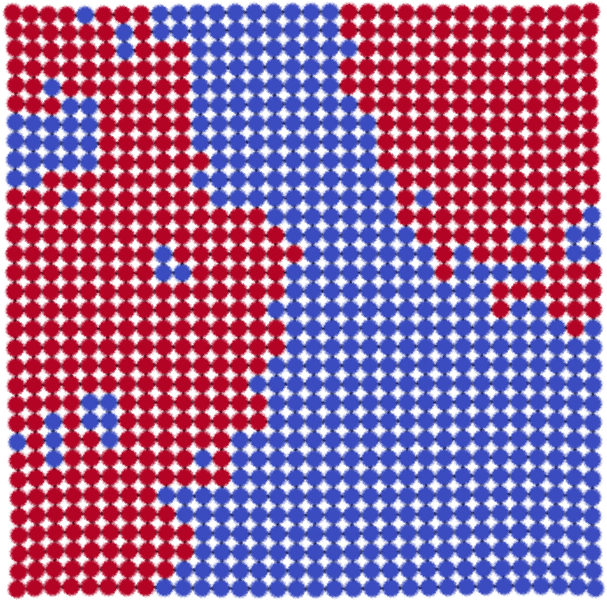}}
  \centerline{(c) label = 1}\medskip
\end{minipage}
\hfill
\begin{minipage}[b]{0.48\linewidth}
  \centering
  \centerline{\includegraphics[width=4.0cm]{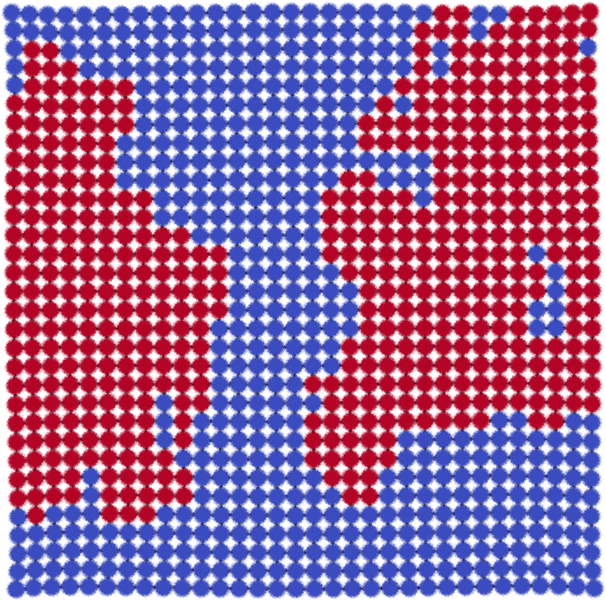}}
  \centerline{(d) label = 1}\medskip
\end{minipage}
\caption{Two negative and two positive examples from 32x32 grid Color-connectivity dataset.}
\label{fig:islands32x32viz}
\end{figure}

\end{document}